\documentclass{bmvc2k}
\usepackage{multirow}
\usepackage{amsmath,amssymb}
\usepackage{color}
\usepackage{ulem}
\title{Reference Guided Image Inpainting \\ using Facial Attributes}
\addauthor{Dongsik Yoon}{kevinds1106@korea.ac.kr}{1}
\addauthor{Jeonggi Kwak}{kjk8557@korea.ac.kr}{1}
\addauthor{Yuanming Li}{lym7499500@korea.ac.kr}{1}
\addauthor{David Han}{dkh42@drexel.edu}{2}
\addauthor{Youngsaeng Jin}{youngsjin@korea.ac.kr}{1}
\addauthor{Hanseok Ko$^*$}{hsko@korea.ac.kr}{1}
\addinstitution{
Korea University\\
Seoul, South Korea
}
\addinstitution{
Drexel University\\
Philadelphia, USA
}
\runninghead{D.Yoon et al.}{Reference Guided Image Inpainting}

\begin{document}
\maketitle
\begin{abstract}
Image inpainting is a technique of completing missing pixels such as occluded region restoration, distracting objects removal, and facial completion. Among these inpainting tasks, facial completion algorithm performs face inpainting according to the user direction. Existing approaches require delicate and well controlled input by the user, thus it is difficult for an average user to provide the guidance sufficiently accurate for the algorithm to generate desired results. To overcome this limitation, we propose an alternative user-guided inpainting architecture that manipulates facial attributes using a single reference image as the guide. Our end-to-end model consists of attribute extractors for accurate reference image attribute transfer and an inpainting model to map the attributes realistically and accurately to generated images. We customize MS-SSIM loss and learnable bidirectional attention maps in which importance structures remain intact even with irregular shaped masks.  Based on our evaluation using the publicly available dataset CelebA-HQ, we demonstrate that the proposed method delivers superior performance compared to some state-of-the-art methods specialized in inpainting tasks.
\end{abstract}
\section{Introduction}
\label{sec:intro}
The recent development of Generative Adversarial Networks (GANs)\cite{17} based image manipulation techniques led to a slew of smartphone applications in the social network scene, as users find it amusing to combine two different facial images into one or other combinations. One variant of this type of manipulation or generation is when an image is given with a mask such that the masked region needs to be filled in appropriately with feature contents from another image. This problem is a special case of inpainting over masked regions with content information taken from another image. 

Image inpainting is one of the techniques in computer vision that handles missing or damaged portions of an image by filling them in with plausible contents. As it is an old problem in computer vision, many methods have been proposed in the past. Among them, patch-based methods\cite{1, 3, 5} complete occluded regions by applying appropriate scene segments based on contextual statistical similarities. These methods perform reasonably when the missing regions are part of backgrounds. However, the method's performance degrades significantly when the occlusions are of a part of an object such as a face. Restoring a face image with occlusion is particularly a difficult challenge for these traditional methods, as the patch used for filling in the missing region may not match well with the rest of the image. Compared to the traditional methods, GANs based inpainting are able to synthesize tractable results in comparison with non-deep learning methods, particularly in facial image restoration. With effective training, these generative models are capable of completing a plausible image while maintaining coherent consistency between the filled-in region and the rest of the image.

Unlike the traditional inpainting of letting the GANs fill in the masked region by itself, the filling process can be guided by a user. Thus, the filled-in region can be manipulated per the user guidance. There have been a good amount of research efforts in manipulating facial regions, such as adding glasses or mustache, using GANs based methods \cite{16, 20, 13, 24, 25, 48}. These methods, however, are based on transforming a complete facial image without any missing parts.
An alternate facial manipulation task is when some regions of the input facial image are masked. In such a task, inpainting is applied to those missing regions based on the user direction. Specifying these directives for useful region manipulation can be difficult. Numerous studies\cite{11, 12, 14, 43} have been suggested to address this issue, using additional conditions such as user specified edges, colors, or landmarks. However, employing these conditions is not straightforward for an average user since accurate specifications are needed for generating user desired images. 

To enable a typical user to manipulate facial images with masked regions, we propose a novel method based on using a reference image as a guide in the inpainting process. Input to our approach consists of an image with a masked region and an intact reference image. Our network first extracts attributes from the reference image so that the inpainting process will fill in the masked region according to the extracted attributes of the reference image. Thus, our architecture consists of an inpainting model and attribute extractors. Our generator adopts Learnable Bidirectional Attention Maps (LBAM)\cite{9} in each layer to allow inpainting of any irregular shaped mask specified by the user. To ensure that the attributes of the generated image closely reflect the reference image attributes, we developed an attributes constraint loss by minimizing the difference between the extracted attributes of the reference image and the fake image. We train this architecture in an end-to-end framework by utilizing CelebA-HQ dataset with their facial attributes labels. Additionally, we adopt Multi-Scale Structure Similarity (MS-SSIM)\cite{27} loss\cite{28} to maintain facial structure consistency.

The main contributions of our work are as follows: (1) We propose a novel framework for an easy-to-use facial image manipulation by using a single reference image as a guide. (2) By applying random attributes, our method is capable of generating pluralistic images while maintaining reference attributes.
(3) The proposed method is capable of inpainting masked regions effectively while delivering manipulated images reflecting closely to the user intentions.

\section{Related Work}
\label{sec:rela}
\subsection{Traditional Inpainting}
Image inpainting has been continuously studied as a method to complete missing regions. Traditional diffusion image inpainting methods used to fill only small, narrow holes such as scratches with surrounding pixels. Patch-based methods\cite{1, 3} propagated information from the background area to the hole using patch similarity as a way to fill larger holes. Patchmatch\cite{5}, which improved the aforementioned methods, was able to synthesize more realistic textures using a fast nearest neighbor field algorithm. But these methods unable to synthesize novel objects that are not in the ground truth.
\subsection{Inpainting by Deep Generative Model}
Recently, learning based GANs methods used for image inpainting. CE\cite{6} was the first inpainting model using a deep neural network, that the mask part of the center square was completed using adversarial loss and $L2$ pixel-wise loss. However, this method was suffered to synthesis a plausible novel object that was not trained. GLCIC\cite{7} used two auxiliary discriminators to solve that suffered to synthesis novel objects of existing methods. CA\cite{18}, one of the pioneer attention-based methods, trained coarse-to-fine networks for image inpainting. PConv\cite{8} was a renowned model that improved the inpainting performance for irregular masks by using partial convolution. Partial convolution\cite{8, 26} was effective to prevent convolution filters capture zeros when passing through the hole. More recently, LBAM\cite{9} proposed learnable bidirectional maps that able to synthesize more realistic inpainting for irregular masks. In contrast to PConv\cite{8}, this model utilized bidirectional attention maps for re-normalization of features on U-net\cite{10} architecture. GConv\cite{21} proposed gated convolution that learning dynamic feature by gating mechanism for each spatial region, it further adopted sketch condition to help user modifying images.

Most image inpainting methods synthesize sole one result for each masked image, even if there more reasonable possibilities. Contrary to the existing method, PIC\cite{22} used short$+$long term attention layer to produce pluralistic results. Additionally, this method addressed a probabilistic principled framework with two parallel paths called reconstructive path and generative path. Pii-GAN\cite{23} used a novel style extractor that able to extract style features from ground truth for input into the generator. This method synthesized a variety of results coherent with the contextual semantics of the input image.

\subsection{Facial Manipulate by Generative Model}
There was a variety of research in manipulating facial regions, such as adding glasses or mustache, using attributes. AttGAN\cite{13}, was a novel approach to control face image by a generative model trained with facial attributes. PA-GAN\cite{16} achieved to disentangle irrelevant attributes using a progressive attention network. While these approaches\cite{13, 16} transform a complete facial image without any missing parts. An alternate facial manipulation task is when some regions of the input facial image are masked. EdgeConnect (EC)\cite{12} utilized edge information for image inpainting, therefore manipulate facial features with guidance edge. LaFIn\cite{14} transformed facial images into different face structures by utilizing landmark information. SC-FEGAN\cite{11} proposed user-guided facial inpainting using various conditions. In this work, users inputted free-form sketch, mask, and color for user desired facial editing. These aforementioned methods\cite{12, 14, 11} are not straightforward for an average user since accurate specifications are needed for generating user desired images.

\begin{figure}
\setlength{\abovecaptionskip}{-10pt}
\begin{center}
   \includegraphics[scale=0.43]{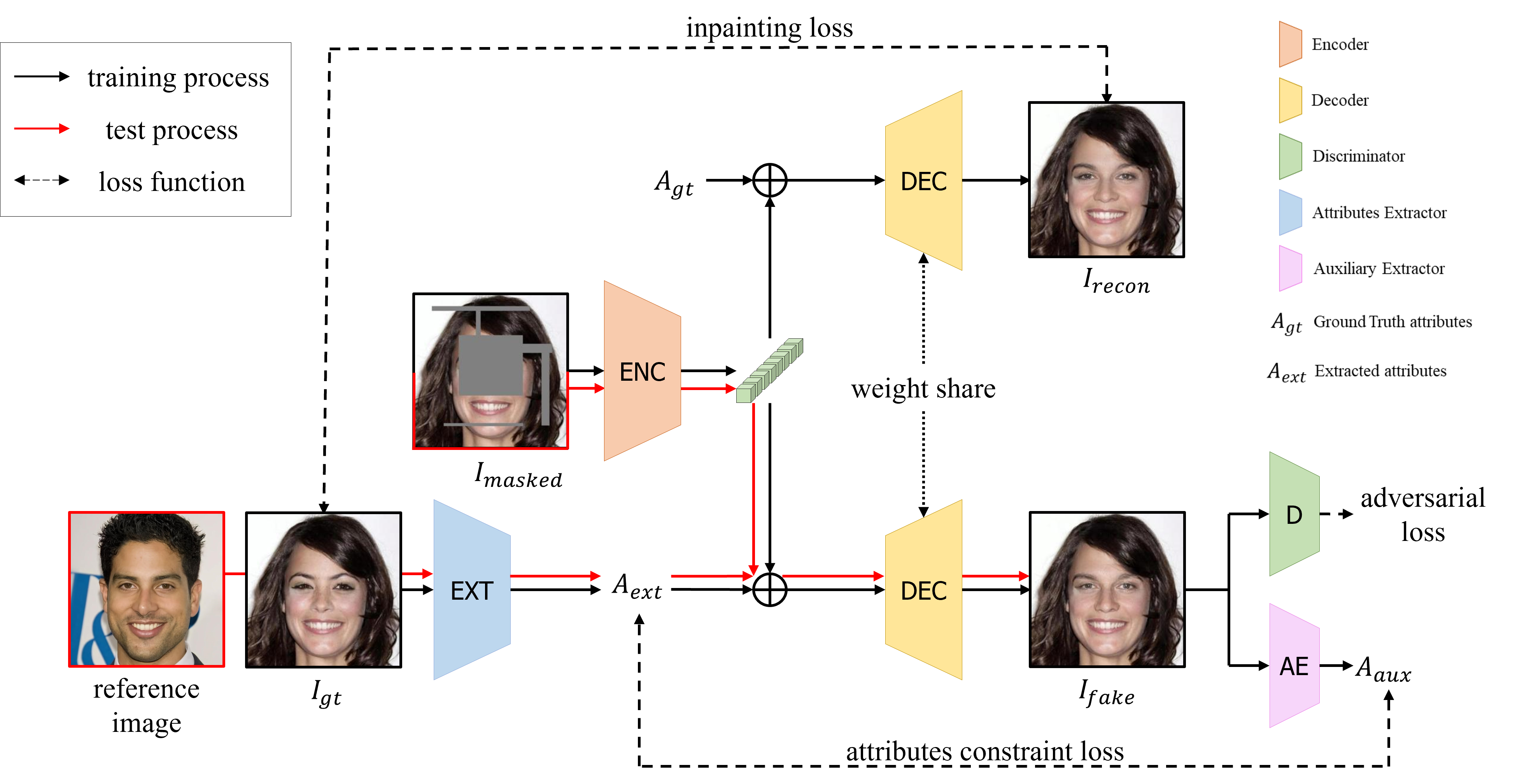}
\caption{Summary of the proposed architecture. $I_{masked}$ and $M$ are the input of $G$, we omit $M$ in this figure to express clearly our framework. For the test stage (red line), the user extract desired attributes using our attributes extractor to a reference image.}
\label{fig1}
\end{center}
\vspace{-5mm}
\end{figure}

\section{Proposed Method}
\label{sec:prop}
\subsection{Model Architecture}
This section introduces our user-guided inpainting architecture. As shown in Figure \ref{fig1}, our architecture consists of four models, generator, discriminator, attributes extractor, and auxiliary extractor. Our generator adopts LBAM\cite{9} in each layer. Since we allow the user to modify random parts of the face, we customize bidirectional attention maps that has shown powerful performance for irregular shaped hole inpainting. Let ${I}_{gt}$ be a ground truth image and its attributes be ${A}_{gt}$. We define the masked image as,
\begin{equation}\label{1}
I_{masked}=I_{gt} \odot M,
\end{equation}
where $M$ is the input mask that occluded portion value as 0. We input $I_{in}=[I_{masked},\,M]$ into our generator $G(\cdot)$, during the training process. We encode $I_{in}$ into latent features and concatenate them with ${A}_{gt}$ to reconstruct the original image. Thus, we decode the combined features to complete the image.
\begin{equation}\label{2}
I_{recon}=G(I_{in},\,A_{gt})
\end{equation}
Moreover, we create another combined feature using extracted attributes $A_{ext}=\text{Ext}(I_{gt})$, where $\text{Ext}(\cdot)$ is the attributes extractor. Then we decode another combined features for edit according to the extracted attributes.
\begin{equation}\label{4}
I_{fake}=G(I_{in},\,A_{ext})
\end{equation}

\begin{figure}[!t]
\setlength{\abovecaptionskip}{-10pt}
\begin{tabular}{@{\,}c@{\,}|@{\,}c@{\,}c@{\,}c@{\,}c@{\,}c@{\,}c@{\,}@{\,}c@{\,}}
\bmvaHangBox{\includegraphics[width=1.7cm]{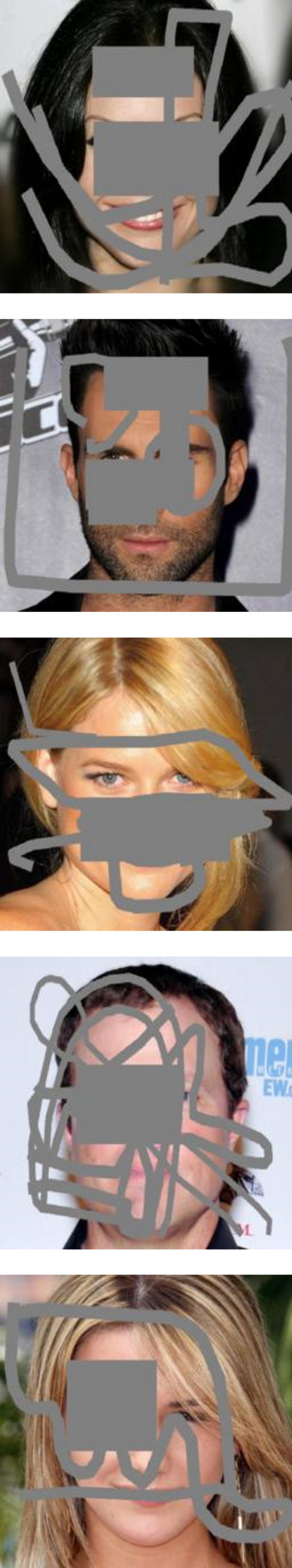}}&
\bmvaHangBox{\includegraphics[width=1.7cm]{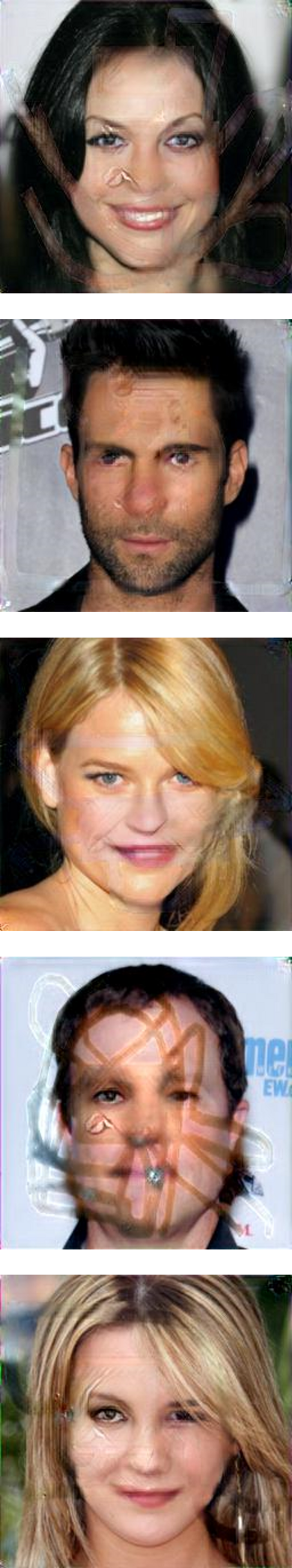}}&
\bmvaHangBox{\includegraphics[width=1.7cm]{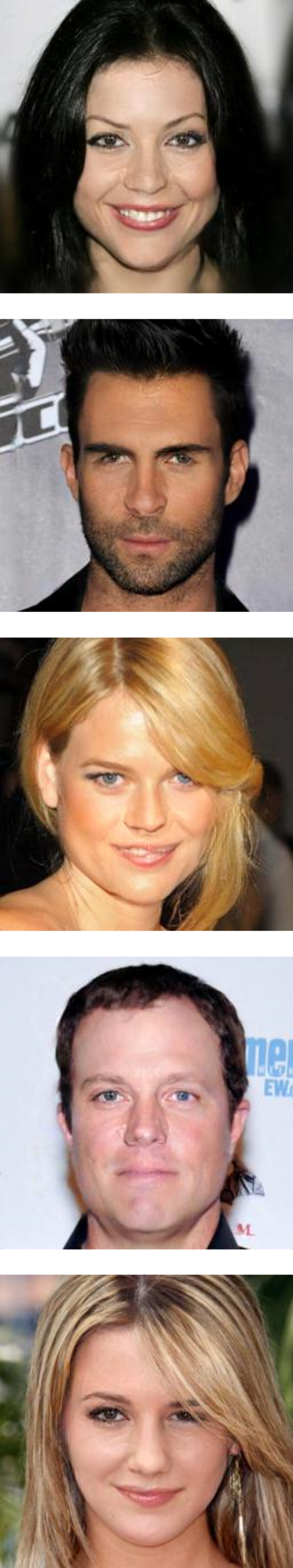}}&
\bmvaHangBox{\includegraphics[width=1.7cm]{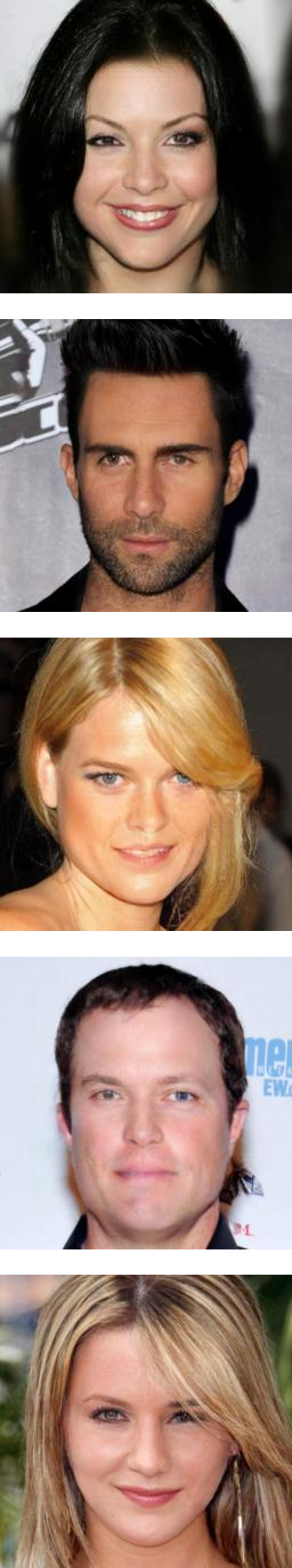}}&
\bmvaHangBox{\includegraphics[width=1.7cm]{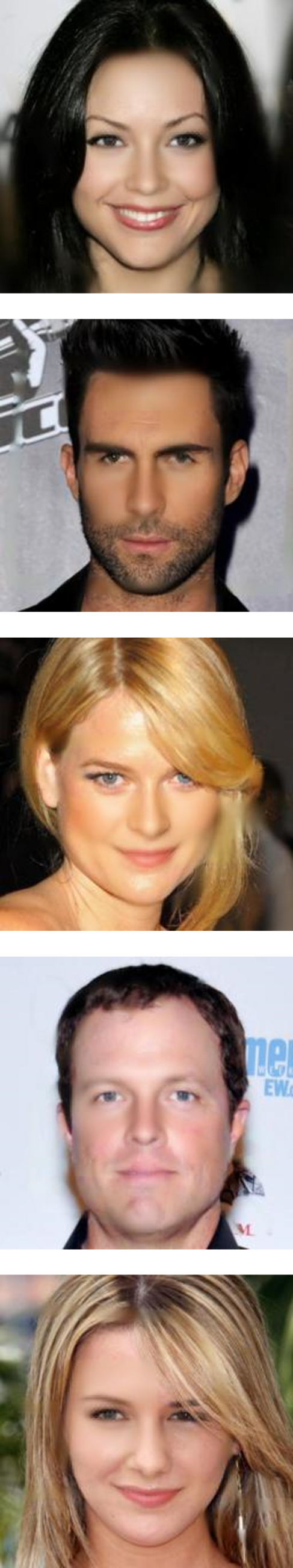}}&
\bmvaHangBox{\includegraphics[width=1.7cm]{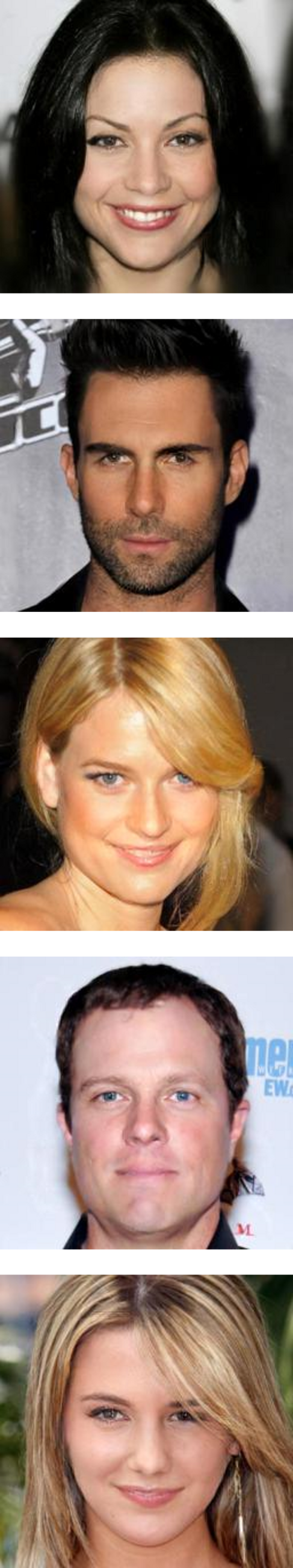}}&
\bmvaHangBox{\includegraphics[width=1.7cm]{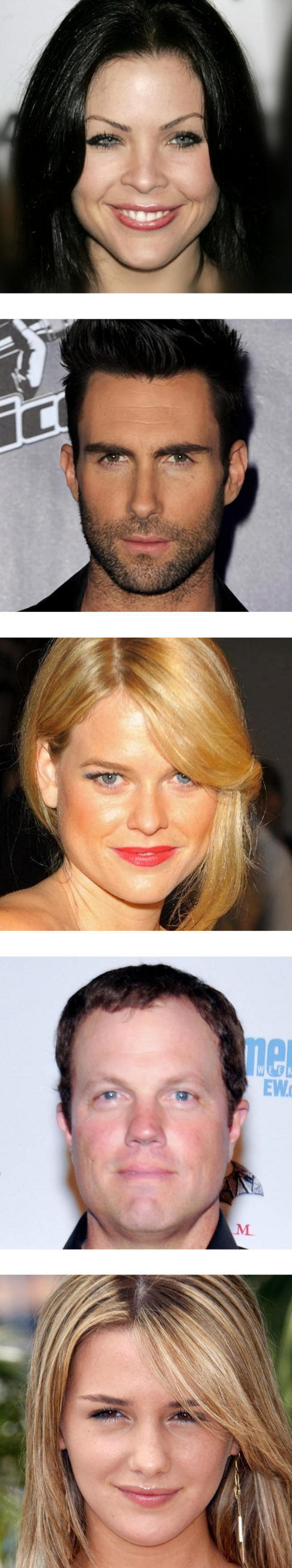}}
\\
Input&PConv\cite{8}&LBAM\cite{9}&EC\cite{12}&MLGN\cite{28}&Ours&Ground Truth
\end{tabular}
\\
\caption{Qualitative comparison with PConv\cite{8}, LBAM\cite{9}, EC\cite{12}, MLGN\cite{28}, and Ours.}
\label{fig2}
\vspace{-2mm}
\end{figure}

In the process of generating $I_{recon}$ and $I_{fake}$, the decoders share their weights. We calculate only inpainting loss to $I_{recon}$ with $I_{gt}$, and we encode $I_{fake}$ into discriminator and auxiliary extractor for calculating adversarial loss and attributes constraint loss, respectively. 
In discriminator, we adopt Spectral Normalization\cite{21, 29} which is fast and stable with a simple formulation in comparison to the other normalization. The attributes extractor predicts the attributes of an input facial image with VGG-16 networks\cite{32}. It is comprised of a feature extracting layer and a fully connected layer for accurate attribute extraction. Our auxiliary extractor $\text{AE}(\cdot)$ is comprised of only convolutional layers to extract attributes of fake images and learn the attributes extractor smoothly. If the attributes extractor aims to train with ${A}_{gt}$ directly, it trains to depend only on fixed ground truth images. Therefore, we design to train an auxiliary extractor by using ${I}_{gt}$ and its attributes before training attributes extractor with ${A}_{aux} = \text{AE}({I}_{fake})$. By doing so, it promotes the attributes extractor for delicate attribute extraction and ensures that the attributes of the fake image closely reflect the reference image attributes.


\begin{figure}[!t]
\subfiglabelskip=1pt
\addtolength{\subfigcapskip}{-4pt}
\begin{center}
\subfigure[Mask]{{\includegraphics[scale=0.4]{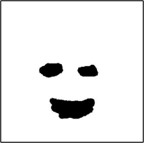} }}%
\subfigure[Ground Truth]{{\includegraphics[scale=0.4]{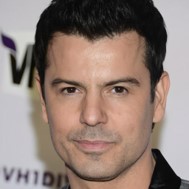} }}%
\subfigure[Input]{{\includegraphics[scale=0.4]{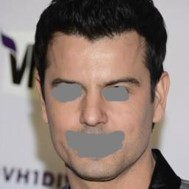} }}%
\subfigure[$\text{PIC}_{1}$\cite{22}]{{\includegraphics[scale=0.4]{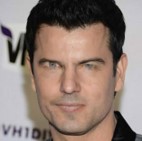} }}%
\subfigure[$\text{PIC}_{2}$]{{\includegraphics[scale=0.4]{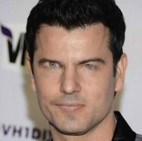} }}%
\subfigure[$\text{PIC}_{3}$]{{\includegraphics[scale=0.4]{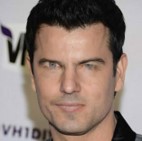} }}%
\vspace{-3mm}
\subfigure[$\text{PD-GAN}_{1}$\cite{40}]{{\includegraphics[scale=0.4]{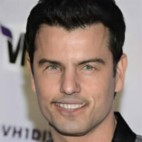} }}%
\subfigure[$\text{PD-GAN}_{2}$]{{\includegraphics[scale=0.4]{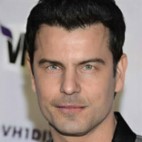} }}%
\subfigure[$\text{PD-GAN}_{3}$]{{\includegraphics[scale=0.4]{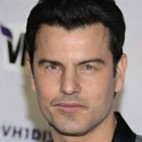} }}%
\subfigure[$\text{Ours}_{1}$]{{\includegraphics[scale=0.4]{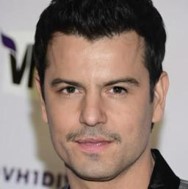} }}%
\subfigure[$\text{Ours}_{2}$]{{\includegraphics[scale=0.4]{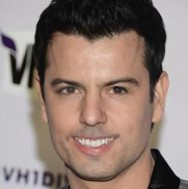} }}%
\subfigure[$\text{Ours}_{3}$]{{\includegraphics[scale=0.4]{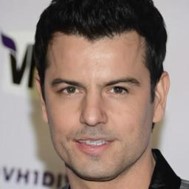} }}%
\vspace{-3mm}
\setcounter{subfigure}{0}
\subfigure[Mask]{{\includegraphics[scale=0.4]{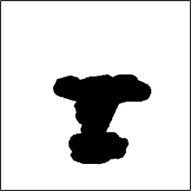} }}%
\subfigure[Ground Truth]{{\includegraphics[scale=0.4]{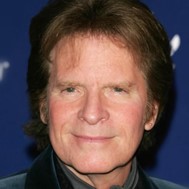} }}%
\subfigure[Input]{{\includegraphics[scale=0.4]{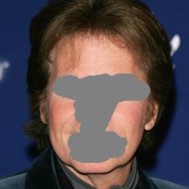} }}%
\subfigure[$\text{PIC}_{1}$\cite{22}]{{\includegraphics[scale=0.4]{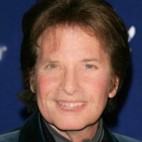} }}%
\subfigure[$\text{PIC}_{2}$]{{\includegraphics[scale=0.4]{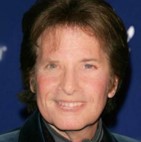} }}%
\subfigure[$\text{PIC}_{3}$]{{\includegraphics[scale=0.4]{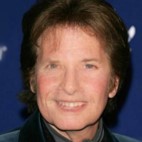} }}%
\vspace{-3mm}
\subfigure[$\text{PD-GAN}_{1}$\cite{40}]{{\includegraphics[scale=0.4]{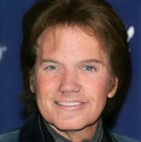} }}%
\subfigure[$\text{PD-GAN}_{2}$]{{\includegraphics[scale=0.4]{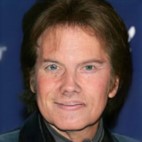} }}%
\subfigure[$\text{PD-GAN}_{3}$]{{\includegraphics[scale=0.4]{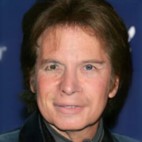} }}%
\subfigure[$\text{Ours}_{1}$]{{\includegraphics[scale=0.4]{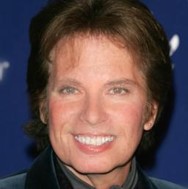} }}%
\subfigure[$\text{Ours}_{2}$]{{\includegraphics[scale=0.4]{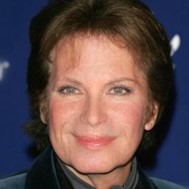} }}%
\subfigure[$\text{Ours}_{3}$]{{\includegraphics[scale=0.4]{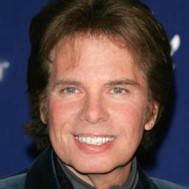} }}%
\caption{Pluralistic qualitative comparison with state-of-the-art methods with PIC\cite{22}, PD-GAN\cite{40}, and Ours. Images other than Ours take from the paper\cite{40}.  We excluded these images from the training process for comparison.}\label{fig3}
\vspace{-5mm}
\end{center}
\end{figure}

\subsection{Loss function}\label{fig3}
In order to deliver manipulated images reflecting closely to the user intention while masked regions are effectively inpainted, we divided our loss into three separate losses: inpainting loss, attributes constraint loss, and adversarial loss.\\
\noindent\textbf{Inpainting Loss.} Our proposed inpainting loss is comprised of reconstruction loss, perceptual loss, style loss, and MS-SSIM loss, for image completion. Reconstruction loss completes erased regions using $l1$-norm error. We calculate hole region and valid region respectively by comparing the generated image with the ground truth.
\begin{gather}\label{5}
\begin{split}
I_{hole}=I_{recon}\odot\,(1-M) \\
I_{valid}=I_{recon}\odot\,M
\end{split}
\end{gather}
\begin{gather}\label{6}
\begin{split}
L_{hole}=||I_{hole}- \, I_{gt}\odot\,(1-M)||_{1} \\
L_{valid}=||I_{valid}- \, I_{masked}||_{1}
\end{split}
\end{gather}

\begin{figure}[!t]
\centering
\setlength{\abovecaptionskip}{-10pt}
\includegraphics[scale=0.3]{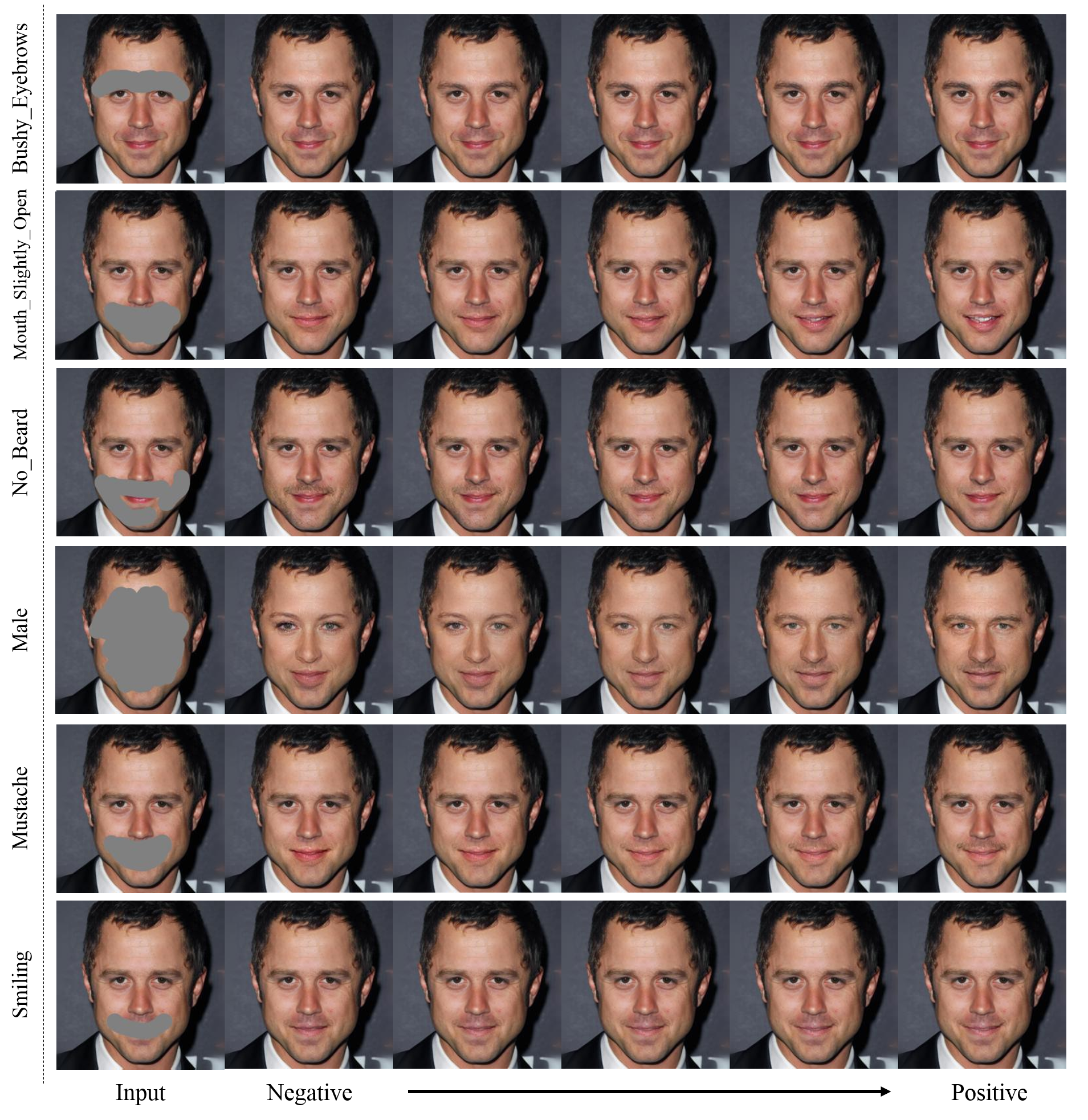}
\caption{Illustration of attributes intensity control. From left to right are: attributes alter gradually negative to positive.}
\vspace{-2mm}
\label{fig4}
\end{figure}

Additionally, we utilize perceptual loss and style loss that most of the existing image inpainting tasks have used. We calculate these losses with the VGG-16 network pre-trained on ImageNet\cite{33}. As the name suggests, perceptual loss measures the distance between the generated image and the feature maps of the ground truth image. It promotes capturing high-level semantics and image quality be consistent with human perception. We defined completion image as  $I_{comp}=I_{masked} \oplus I_{hole}$, and defined perceptual loss as
\begin{equation}\label{7}
L_{percep}=\frac{1}{N}\sum_{i=1}^{N}||\phi_{i}(I_{gt})-\phi_{i}(I_{comp})||^{2},
\end{equation}
where $\phi_{i}$ is the feature maps of the $i\,'th$ layer of a pre-trained network and $N$ is the number of layers in VGG-16 network. We adopt style loss, as defined by \cite{34}, which alleviates "checkerboard" artifacts due to transposed convolution layers. The style loss is defined as
\begin{equation}\label{8}
L_{style}=\frac{1}{N}\sum_{j=1}^{N}||G^{\phi}_{j}(I_{gt})-G^{\phi}_{j}(I_{comp})||^{2},
\end{equation}
where $G^{\phi}_{j}$ is a gram matrix comprised from feature maps ${\phi}_{j}$. 
Furthermore, we customize another loss using MS-SSIM\cite{27}, which is one of the methods to compare image quality. MS-SSIM loss is utilized at facial image inpainting tasks\cite{28}, and it preserves important facial structure information such as a nose or a mouth.
\begin{equation}\label{9}
L_{\text{MS-SSIM}}=1-\frac{1}{N}\sum_{i=1}^{N}\text{MS-SSIM}_{n}
\end{equation}
\noindent\textbf{Attributes Constraint Loss.} To ensure that the attributes of the generated image closely reflects the user desired attributes, we developed an attributes constraint loss. We minimize the difference between the extracted attributes of the reference image and the fake image to make the model respond more sensitively to input attributes. Moreover, we minimize the additional difference between $\text{AE}({I}_{gt})$ and ${A}_{gt}$ in order to train the auxiliary extractor. Unlike other works that perform facial manipulation using classifier \cite{13, 16}, our constraint loss induces accurate extraction of attributes.
\begin{equation}\label{10}
L_{attr}=\text{MSE}(A_{aux},\,A_{ext})+\text{MSE}({A}_{gt},\text{AE}({I}_{gt}))
\end{equation}
\noindent\textbf{Adversarial Loss.} We adopt WGAN-GP\cite{31} that optimizes the Wasserstein distance by further employing gradient-penalty to calculate the adversarial loss. Specifically, it utilizes the Earth-Mover distance to compare synthesized and real distributions of high-dimensional data. Following this approach, our adversarial loss $L_{G}$ and $L_{D}$ are denoted as,

\begin{gather}
 L_{G}=\mathbb{E}_{I_{in},\,A_{ext}}[D(G(I_{in},\,A_{ext}))], \\
 L_{D}=\mathbb{E}_{I_{gt}}[D(I_{gt})]-\mathbb{E}_{I_{fake}}[D(I_{fake})]-\lambda_{gp}\mathbb{E}_{\hat{I}}[(||	\nabla_{\hat{I}}D(\hat{I})||_{2}-1)^{2}].
\end{gather}

Overall our loss denoted as,
\begin{equation}\label{13}
L_{all}=\lambda_{adv}L_{adv}+L_{attr}+\lambda_{ssim}L_{\text{MS-SSIM}}+\lambda_{sty}L_{style}+\lambda_{per}L_{percep}+\lambda_{hole}L_{hole}+L_{valid},
\end{equation}
where, $\lambda$ are hyper-parameters that regulate the relative importance of the terms.

\begin{figure*}[!t]
\centering
\setlength{\abovecaptionskip}{-10pt}
\includegraphics[scale=0.235]{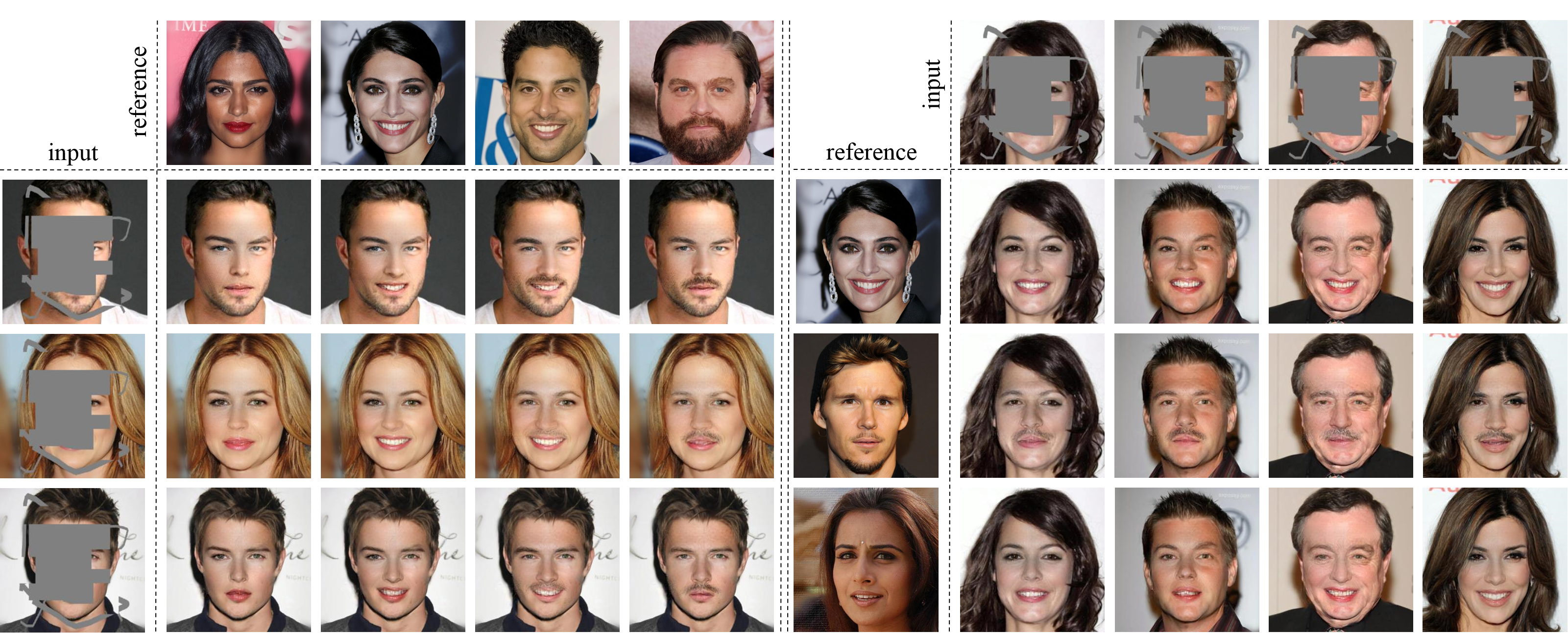}
\caption{Overview of proposed facial image inpainting.}
\vspace{-2mm}
\label{fig5}
\end{figure*}

\section{Experiments}
\label{sec:expri}
\subsection{Implement Details}
We used the Pytorch library for implementation. Our hyper-parameters $\lambda_{adv}$, $\lambda_{ssim}$, $\lambda_{sty}$, $\lambda_{per}$, and $\lambda_{hole}$ are set to 0.1, 3, 120, 0.01, and 6 respectively. These hyper-parameters were adjusted based on the qualitative performance of the empirical train process with reference studies\cite{9, 28}. Our models were optimized using Adam optimizer\cite{39}.
We evaluate all the models in this paper using CelebA-HQ dataset\cite{35} and utilize 28,000 selected images for training to optimize parameters and 2000 for testing. Eight attributes with visible impact are chosen in our experiments, including "Bushy$\_$Eyebrows", "Mouth$\_$Slightly$\_$Open", "Big$\_$Lips", "Male", "Mustache", "Smiling", "Wearing$\_$Lipstick", and "No$\_$Beard". For experiments, we trained and evaluated the proposed methods by 256$\times$256 images with irregular holes. In addition, to adopt irregular holes on images, we utilize the Quickdraw irregular mask dataset\cite{36} combined with 85$\times$85 square holes at a random position. Combining square holes with Quickdraw dataset promotes the model to be more robust to irregular holes.

\subsection{Qualitative Comparisons}
First, we compare an image inpainting quality against four state-of-the-art methods and our baseline. Figure \ref{fig2} shows images generated by the proposed method with those generated by the other methods. Images generated by PConv\cite{8} have failed to maintain a facial structure. Our model performed better than all the others in terms of the generated image quality and plausibility. In Figure \ref{fig2}, we adopt the ground truth attributes to only evaluate the quality of the image inpainting task of our model. In our approach, despite the large irregular holes, facial structures  are well preserved. Figure \ref{fig3} compares pluralistic images generated by PIC\cite{22}, PD-GAN\cite{40}, and ours. Compared to the shown existing methods, our method accomplished more believable and diverse instances using random attributes.
Figure \ref{fig4} shows images generated by manipulating each attribute from the positive and to the negative extremes. As shown in Figure \ref{fig4} \ref{fig5}, our method successfully delivers plausible images according to the given mask shapes and attributes of varying degrees.

\begin{table}[!t]
\begin{center}
\begin{tabular}{l|c|c|c|c|c|c}
                                           & Mask      & \multicolumn{1}{c|}{Ours} & \multicolumn{1}{c|}{LBAM} & \multicolumn{1}{c|}{PConv} & \multicolumn{1}{c|}{EC} & \multicolumn{1}{c}{MLGN} \\ \hline\hline
\multicolumn{1}{c|}{\multirow{5}{*}{SSIM}} & Quickdraw & \textbf{0.833} & 0.821          & 0.772 & 0.817 & 0.832          \\ \cline{2-7} 
\multicolumn{1}{c|}{}                      & 10-20\%   & 0.811          & 0.814          & 0.789 & 0.827 & \textbf{0.839} \\ \cline{2-7} 
\multicolumn{1}{c|}{}                      & 20-30\%   & 0.740          & 0.744          & 0.700 & 0.761 & \textbf{0.777} \\ \cline{2-7} 
\multicolumn{1}{c|}{}                      & 30-40\%   & 0.660          & 0.667          & 0.602 & 0.681 & \textbf{0.706} \\ \cline{2-7} 
\multicolumn{1}{c|}{}                      & 40-50\%   & 0.571          & 0.583          & 0.502 & 0.595 & \textbf{0.624} \\ \hline\hline
\multirow{5}{*}{LPIPS}                     & Quickdraw & \textbf{0.042} & 0.047          & 0.088 & 0.047 & 0.063          \\ \cline{2-7} 
                                           & 10-20\%   & 0.068          & 0.057 & 0.080 & \textbf{0.051} & 0.065          \\ \cline{2-7} 
                                           & 20-30\%   & 0.103          & 0.087 & 0.130 & \textbf{0.082} & 0.103          \\ \cline{2-7} 
                                           & 30-40\%   & 0.146          & \textbf{0.126} & 0.214 & 0.128 & 0.148          \\ \cline{2-7} 
                                           & 40-50\%   & 0.198          & \textbf{0.172} & 0.309 & 0.183 & 0.201          \\ \hline\hline
\multirow{5}{*}{FID}                       & Quickdraw & \textbf{24.91} & 25.79          & 33.69 & 27.49 & 28.45          \\ \cline{2-7} 
                                           & 10-20\%   & 28.53          & 27.63          & 31.53 & \textbf{25.65} & 26.73 \\ \cline{2-7} 
                                           & 20-30\%   & 35.26 & 36.51          & 61.03 & \textbf{34.80} & 38.34          \\ \cline{2-7} 
                                           & 30-40\%   & \textbf{43.29} & 48.47          & 127.2 & 47.14 & 52.54          \\ \cline{2-7} 
                                           & 40-50\%   & \textbf{56.67} & 64.40          & 207.3 & 63.75 & 73.07          \\ \hline
\end{tabular}
\end{center}
\caption{Quantitative comparison on CelebA-HQ. The best results of each row is boldfaced.}\label{table1}
\end{table}

\subsection{Quantitative Comparisons}
We compare an image inpainting quantitatively against four existing methods and ours with different types and sizes of masks. Despite existing methods specialize in only image inpainting tasks, as shown in Table \ref{table1}, our method outperforms similarly or superiorly over the three commonly used metrics SSIM, LPIPS\cite{42}, and FID\cite{41}. We demonstrate that our method edits according to input attributes, concurrently while maintains high inpainting performance.

\subsection{Ablation Study}
\begin{figure*}[!t]
\centering
\includegraphics[scale=0.4]{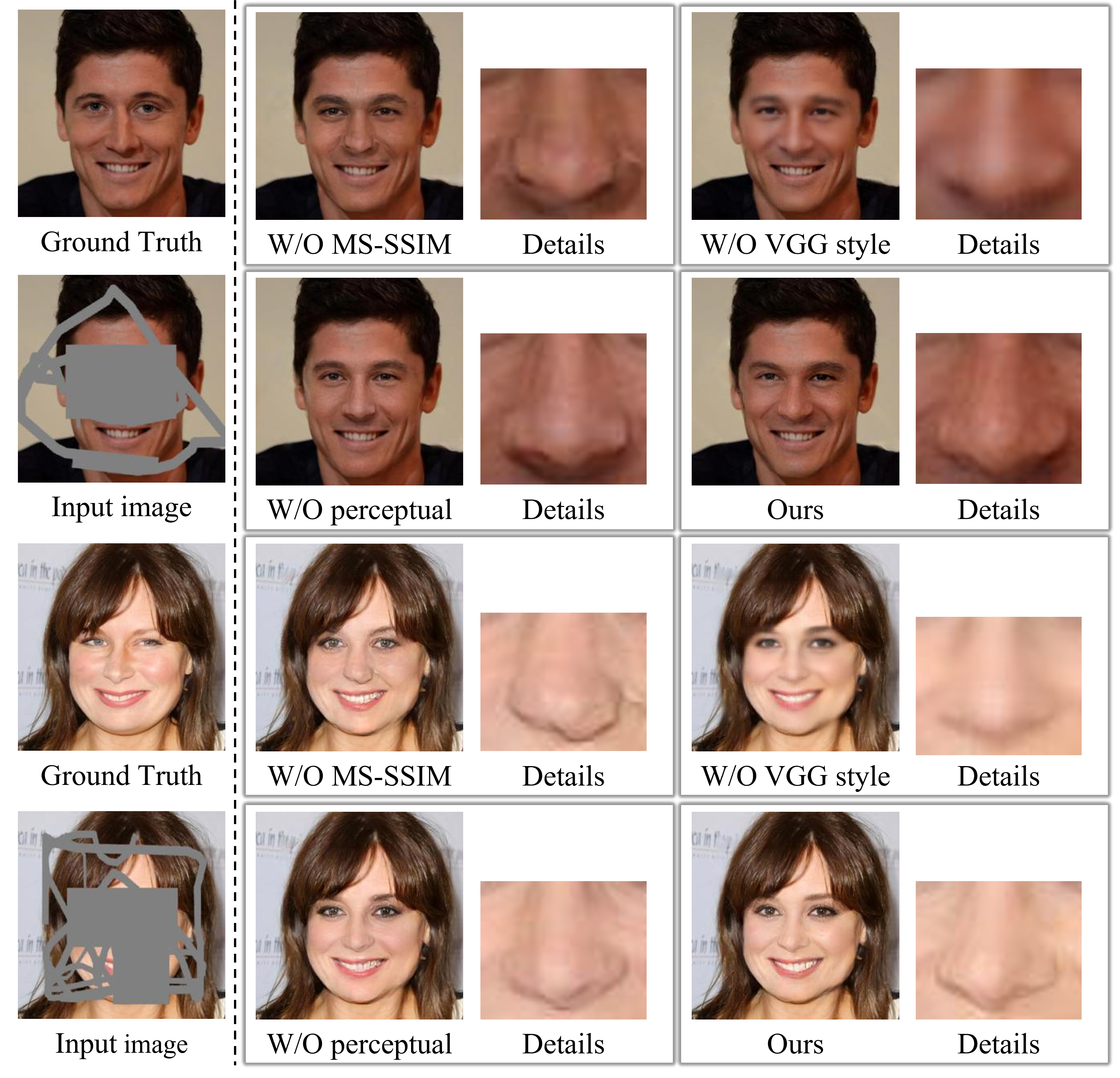}
\caption{Qualitative comparisons with and without specific loss terms.}
\label{fig6}
\vspace{-2mm}
\end{figure*}

To justify and validate the effectiveness of the proposed loss function, we conduct qualitative comparisons. Figure \ref{fig6} shows the inpainting results with and without specific loss terms. By using perceptual loss and style loss the model learns to produce texture details in the output without any distortions or blurriness.  Particularly, we demonstrate that MS-SSIM loss is a key component of our framework to maintain facial structure.

\section{Conclusion}
\label{sec:conc}
In this paper, we proposed a high-quality user-guided inpainting architecture that manipulates facial attributes of a masked image by injecting the attributes from another intact reference image. The proposed novel architecture combines LBAM with an attributes extractor to reflect the features from a reference image chosen by the user. While applying a reference image as the guide for image manipulation may not empower the user with arbitrary control, choice of the reference image is arbitrary by the user and the user guidance on image manipulation becomes much simpler. Experimental results demonstrated that our method delivers high inpainting performance while at the same time making it much easier for a user to guide the inpainting. Further, we generated multiple and diverse plausible images for a single masked input from the extracted attributes. In future work, we aim to user-guided facial inpainting for non-annotated datasets using latent features of a reference image, not facial attributes.
\clearpage

\section*{Acknowledgment}
This research was supported by Deep Machine Lab (Q2109331).

\bibliography{egbib} 
\end{document}